# Report on the First Knowledge Graph Reasoning Challenge 2018
– Toward the eXplainable AI System –

Takahiro Kawamura[1][0000−0002−2765−6232], Shusaku Egami[2], Koutarou Tamura[3,4], Yasunori Hokazono[4], Takanori Ugai[5], Yusuke Koyanagi[5], Fumihito Nishino[5], Seiji Okajima[5], Katsuhiko Murakami[5], Kunihiko Takamatsu[6], Aoi Sugiura[7], Shun Shiramatsu[8], Shawn Zhang[8], and Kouji Kozaki[9]

1 National Agriculture and Food Research Organization, Japan
2 National Institute of Maritime, Port and Aviation Technology, Japan
3 NRI digital, Ltd.
4 Nomura Research Institute, Ltd.
5 Fujitsu Laboratories Ltd.
6 Kobe Tokiwa University
7 Kobe City Nishi-Kobe Medical Center
8 Nagoya Institute of Technology
9 Osaka Electro-Communication University

**Abstract.** A new challenge for knowledge graph reasoning started in 2018. Deep learning has promoted the application of artificial intelligence (AI) techniques to a wide variety of social problems. Accordingly, being able to explain the reason for an AI decision is becoming important to ensure the secure and safe use of AI techniques. Thus, we, the Special Interest Group on Semantic Web and Ontology of the Japanese Society for AI, organized a challenge calling for techniques that reason and/or estimate which characters are criminals while providing a reasonable explanation based on an open knowledge graph of a well-known Sherlock Holmes mystery story. This paper presents a summary report of the first challenge held in 2018, including the knowledge graph construction, the techniques proposed for reasoning and/or estimation, the evaluation metrics, and the results. The first prize went to an approach that formalized the problem as a constraint satisfaction problem and solved it using a lightweight formal method; the second prize went to an approach that used SPARQL and rules; the best resource prize went to a submission that constructed word embedding of characters from all sentences of Sherlock Holmes novels; and the best idea prize went to a discussion multi-agents model. We conclude this paper with the plans and issues for the next challenge in 2019.

**Keywords:** Knowledge graph · Open data · Reasoning · Machine learning

## 1    Background and Goal of the Challenge

In the near future, social systems regarding transportation and the economy are expected to incorporate artificial intelligence (AI)-related techniques. These systems will make critical decisions without human intervention. To safely and securely use AI techniques in society, we need to confirm whether the AI system is working appropriately. However, recent machine learning techniques, such as deep learning, hide the internal process of their decisions and predictions, and humans do not understand the reason for their conclusions. Therefore, there is increasing consideration of AI techniques that have explainability or interpretability, which means the AI system can explain the reason for the conclusion.

Moreover, although machine learning is attracting attention, we believe that the combination of inductive learning techniques and deductive knowledge reasoning techniques will be necessary in the near future. For example, considering the autonomous vehicle, in which the system inevitably has accountability when an incident or accident occurs, there is a strong demand for learning and estimating techniques for situation recognition around the vehicle based on a front camera and radar sensors. However, traffic rules and vehicle operation are predefined knowledge; thus, both must be integrated into the final system to realize the autopilot.

To the best of our knowledge, however, there is no dataset that can be used to evaluate the inductive learning techniques and the deductive knowledge reasoning techniques. Most datasets frequently used for relation estimation, such as FB15k and WN18, only include simple facts, such as hasSpouse and is-a. Such datasets cannot be used for tasks like the autopilot, in which several subtasks must be combined to achieve the overall goal. However, most knowledge bases depend on certain domains like factories[1] and few knowledge bases can be applied as large-scale general datasets for machine learning. The datasets for evaluating explainable AI using estimation and reasoning must include not only the basic relationships of data fragments, but also the temporal, causal, and statistical relationships in real society.

Therefore, we decided to establish a challenge to promote the development of explainable AI, with particular emphasis on combining estimation and reasoning. The goals of this challenge are to **(1)** construct large-scale knowledge graphs including structural and/or sequential relations, such as social problems and human relationships, as common datasets to evaluate reasoning and estimation techniques, **(2)** aggregate such techniques from a wide variety of researchers and information technology (IT) engineers as a means of open science by opening the knowledge graphs to the public, and **(3)** conduct objective evaluation and classification after designing appropriate metrics for explainability. In the challenge, we also used the knowledge graph in a Resource Description Framework (RDF) to provide a dataset and relationships of the data fragments in a unified machine-readable form for estimation and reasoning.

The remainder of this paper is organized as follows. Section 2 introduces the knowledge graph constructed for the 2018 challenge, and Section 3 presents the winning proposed approaches. Section 4 describes the evaluation methods and results, and Section 5 introduces related works. Finally, Section 6 considers the 2019 challenge.

## 2   Knowledge Graph Construction

The subject of the first challenge in 2018 was *The Speckled Band*, a short Sherlock Holmes story[1]. The challenge task was to correctly identify the criminals and to explain the reason, such as the evidence and methods, based on a knowledge graph, which represented the case, background, and characters of the novel. The reasons for using a mystery story as the subject are as follows:

– The story includes complicated relationships in real society, but is a virtually closed world; thus, we can set an answer and control several conditions to get this answer.
– Some stories would require statistical processes and machine learning to handle uncertain information and evidence photographs, and some stories would need to complement common sense knowledge not written in the novel; thus, we can promote the integration of estimation and reasoning.
– The story intrinsically has explainability for humans since, if the readers are not convinced, it does not hold as a mystery story.
– Using a famous story could attract attention to the challenge.

### 2.1   Process of holding the challenge

We first held five open workshops from November 2017 to April 2018, where we discussed the schema design of the graph and methodology of the knowledge construction, and we then worked towards the actual construction. The total number of participants in the workshops was 110.

In the schema design, we first discussed the contents to be included in the knowledge for estimation and reasoning, and their expression through the trial construction of the knowledge graph. Following feedback and opinions from the participants, we decided on the basic policy to focus on scenes in a novel and the relationship of those scenes, including the characters, objects, places, etc., with related scenes. In the schema, a scene ID (IRI) has subjects, verbs, objects, etc. as edges to mainly represent five Ws (When, Where, Who, What, and Why). Thus, by tracing the scene IDs, we can query the temporal transition and causal relationship of events and the characters' actions in conjunction with the static information of the characters, objects, and places. Moreover, commonsense knowledge can be added as axioms and rules. Tables such as timetables and pictures, such as evidence photographs, can also be linked to the graph. In addition, sentences in the novel are included as literal values for natural language processing.

---

[1] https://en.wikipedia.org/wiki/Canon_of_Sherlock_Holmes

In the knowledge graph construction, we first extracted sentences to be represented in the graph (approximately 500 sentences in *The Speckled Band*). Participants manually modified the original sentences to the simple sentences and annotated the semantic roles (five Ws in this case) to each clause in the schema defined files in Google Sheets. We then normalized the notations of the subject, verb, object, etc. and added the relationships of the scenes, such as temporal transitions. Finally, we transformed the sheet into a Resource Description Framework (RDF) file. Thus, the knowledge graph includes facts written in the story, testimonies of characters, and the contents introduced by Holmes's reasoning, which are all types of information used to identify criminals. Notably, indirect information not useful for criminal identification, such as emotional landscapes, should also be incorporated in the knowledge graph. However, we leave this as an issue for after the first year.

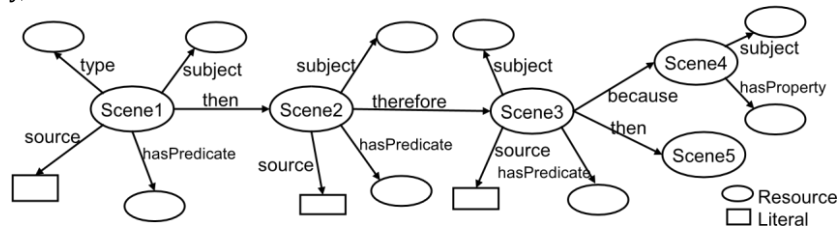

**Fig.1.** Architecture of the knowledge graph

We then opened the knowledge graph to the public and collected the methods to identify criminals and the results. Application guidelines were published at the official website[2] (in Japanese). After opening the knowledge graph to the public, we held three orientation meetings in August, September, and October 2018, and more than 200 participants, including engineers in tech ventures and researchers at universities and companies, conducted the active discussion. The application deadline was the end of October 2018, and presentations of all applications and an awards ceremony were held at an event collocated with the 8th Joint International Semantic Technology Conference (JIST 2018).

### 2.2   Details of the schema

Figure 1 presents the overall architecture of the knowledge graph.

Several properties are used to describe each scene. To summarize the information related to each scene, the properties share a scene ID as the *subject*; noting that the *subject* is not the subject of each scene or sentence. These properties are as follows:

- subject: person or object that is a subject in a scene
- hasPredicate: verb to the above subject in the scene

---

[2] http://challenge.knowledge-graph.jp/

- hasProperty: property to the above subject (a scene has either hasPredicate or hasProperty)
- whom, what, where, how, why: persons, objects, or place that are details of the scene
- when, then, after, if, because, etc.: the relationship between scenes (the values are scene IDs)
- time: absolute time the scene occurs (xsd:DateTime)
- source: original sentences that describe the scene (Literal in English and Japanese)

Figure 2 presents an example of a scene description. If a scene describes a remark from X (rdf:Type of the scene ID is Statement) or a thought of X (rdf:Type of the scene ID is Thought), the kgcc:infoSource property has X as an information source. In addition, there is the case where one or some of subject, whom, what, place, etc. has more than one value in a scene. In the case that the relationship of the values is AND, the scene has more than one triple with the same property and different values. In the case of OR, the value is an instance (resource) of a type ORobj that represents an OR combination, and the resource has several values through the kgcc:orTarget property. To handle the negation form of verbs, the negative verbs are linked to the corresponding positive verbs with the kgcc:Not or kgcc:canNot property. Fig. 3 shows a list of classes and properties, and the knowledge graph that was opened to the public can be accessed at our Linked Data visualization tool[12].

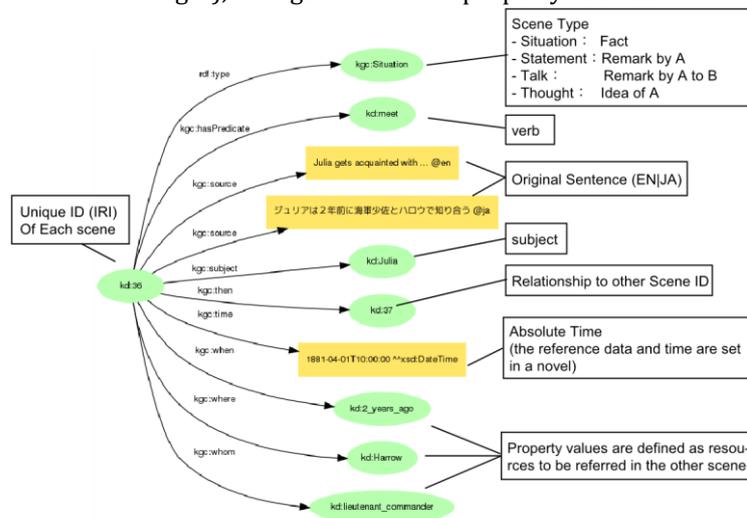

Fig. 2. Example of a scene graph

6     T. Kawamura, S. Egami, K. Kozaki, et al.

## 3     Approach for estimation and reasoning techniques

Estimation and reasoning techniques include several existing information retrieval, relation extraction, and knowledge management techniques; however, since this challenge focuses on explainability for human use in social problems, we note the following unique difficulties. In real social problems, every single issue is an individual case, and it is hard to collect problems with the same pattern. Thus, data and/or knowledge of the problems cannot necessarily be "big data." One single relation estimation, i.e., a reasoning step that can be explained with similarity in embedded vector spaces is insufficient. Aggregating them to lead to the overall goal is essential.

   With respect to the above difficulties, we had eight original submissions (5 submissions with implementation, 3 submissions of ideas only). The following subsections briefly introduce the four approaches that respectively obtained the first prize, the second prize, the best resource prize, and the best idea prize. More details can be found in the results announcement page on the official website.[12]

[12] http://knowledge-graph.jp/visualization/

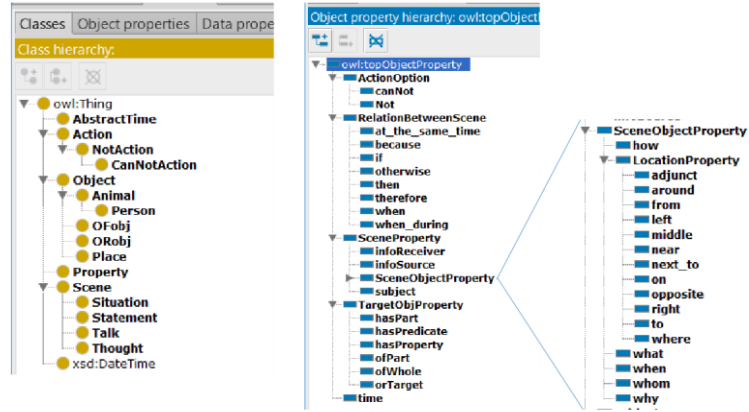

Fig. 3. Class and Property list

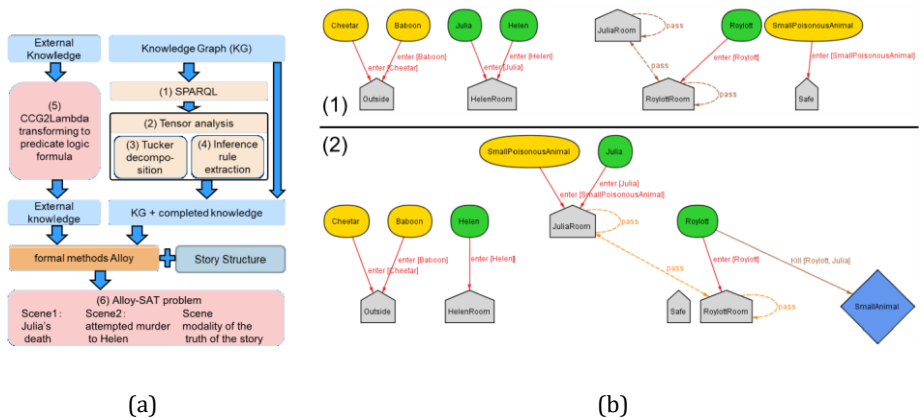

(a)                                              (b)

**Fig.4.** (a)NRI team's approach, (b)One of the solution corresponding to the day of Julia's death: night (1) and mid-night (2). Due to lack of the facts it cannot identify whether the small poisonous animal stays in the safe or Roylott's room in the night.

### 3.1    Submission of NRI

In describing the crime situation mathematically from the knowledge graph in the text, and complementing the missing knowledge, we analyzed the knowledge graph as relational data and considered treating them as satisfiability (SAT) problems while searching for possible situations. An overview of our method is shown in Fig.4(a).

In the following, we explain the contributed methods. For detailed scripts and condition settings, please refer to the materials listed on the official website's results announcement page.

First, we try to complement the knowledge lacking in the information in the text by Tucker decomposition. One scene is described by three axis, SVO forms (subject, predicate, object) to shorten the calculation time. Although some of the complemented knowledge is difficult to interpret, there are a large number of knowledge items suggesting that Animal has done something to Julia, although the accuracy is not high. Among the information on the deceased Julia, the few items on the activities aimed at her were judged to be strong.

Second, we rewrite a given sentence as a predicate logic expression. We analyzed the SAT problem using the Analyzer with the converted predicate logic equation and facts (situation in the scene). Also, we introduced external knowledge such as the list of killing methods for the closed-room murder, structure of the house, and list of animals living in India [3]. Information on actors, killing methods, and buildings / locations common to the scene is represented as a predicate logical expression, and we listed the possibility of the criminal situation shown in Fig.4(b) with the facts on the person and the situation in the scenes "day of Julia's death" and "day of attempted murder to Helen."

Given the testimony and information, Roylott is shown as the perpetrator on both the day of Julia's death and on the day of attempted murder to Helen.

In this experiment, we set the situation in tensor analysis and SAT problem and used the former to complement the information and the latter to search for the matching situation. Only the SVO axis was used for the tensor analysis, whereby it was possible to obtain a suggestion that people do not notice. However, the large number of complementary knowledge items presented is an issue in that it was necessary for people to interpret and select the items.

And, using SAT problem solution has been suggested that mechanical reasoning can be performed while persuasive explanation is retained. In addition, Our method has room for the development to solve with contradict facts by weighted-SAT, which solves a problem by weighting formulas and facts, and to narrow the solution space using evolutionary computation by proposing hypothesis. Upgrading the algorithm of by these approach within the framework of SAT problem is a subject for future works. By using the SAT problem solution method here, we hope that we will help explore basic artificial intelligence and machine learning technology with a high interpretability in the future.

CCG2lambda [4] and AlloyAnalyzer greatly contributed to this analysis.

### 3.2  Submission of Team Kamikotanaka 411, Fujitsu Overall Structure and Created Ontology

Overall Structure Team Kamikotanaka 411 tried to find the criminal by inferring who had the motive, the opportunity, and the method.

An ontology representing motives and methods of murder like Fig 5 was made, and it was added to the provided knowledge graph of "The Adventure of the Speckled Band". We applied programs to judge the motive, opportunity, and method to the extended knowledge graph. The final overall judgment was made manually based on the judgment result of the characters.

Created Ontology: In order to infer the criminal, various kinds of evidence and various analyses are conducted by comparing them with various knowledge related to the crime. This time, the knowledge for the three analyses from a) analysis of motives, b) analysis of opportunities, c) analysis of means. The Criminal White Paper contains statistics on motives for crimes, and we listed the possible motives for the crime mainly from the statistics. Human relations are described based on "An Agent Relationship Ontology" from the viewpoint of describing inheritance. As for the means, since there are various Japanese words which indicate the means of murder, various ways of killing were enumerated on the basis of the backward coincidence of "killing" in the Japanese dictionary. These can be described in detail as structured data with attributes (service, action, place, object) and values.

**Finding crimes from Motives, Opportunities, and Means**

Motives: We set a basic policy that one will have a motive for murder if the target of murder is clear under circumstances where it is not unnatural to have a motive for murder. Then we established a rule to infer the characters concerned through creating an ontology of motives for murder to describe situations in which a motive for murder occurs.

As a result of inference using the processing system of SHACL, the following three results were obtained. 1) *Roylott* may kill *Julia* and *Helen* for *money*, 2) *Villagers* may kill *Roylott* for *self-defense*, 3) *Helen* may kill *Roylott* for *selfdefense*.

Oppotunities: We set a basic policy that any person who could get into the room of Julia on the night of the murder can be the criminal. And the inferring process was divided into a part which infers the relation of time by inferring each whereabouts at the time of the incident and a part which infers the relation of space.

The locations of the characters at the time of the incident were deduced by obtaining the information of the scenes with the same time as the time of the incident, and excluding the scenes which were inferred to be after the occurrence of the incident by the property "then". It is inferred that the five characters who were near the crime scene were as follows: Julia, Helen and Roylott were in their own bedrooms and Roma was in the garden.

Next, we infer whether Helen, Roylott and Roma, other than the murdered Julia, could move to the bedroom where Julia was in. Enumerating the con-nections that are made by the hole, and describing the connections that people cannot pass through.

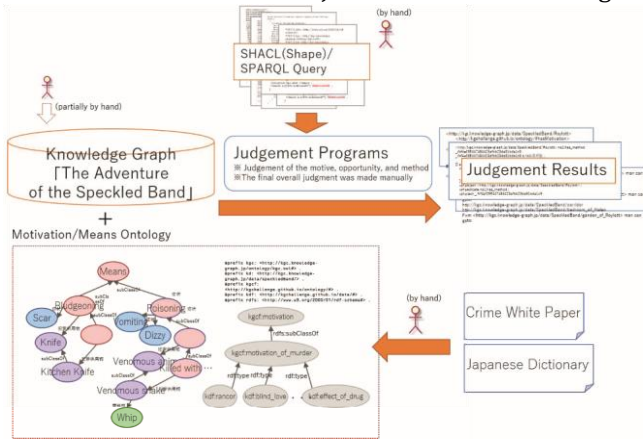

**Fig.5.** Overall Structure

Means: A part which narrows down the killing method based on the condition of the victim and the scene on the night of the incident and a part which deduces the person who satisfies the necessary condition for carrying out the narrowed killing method were implemented. It was inferred from this query that the method of killing was poisoning, and the symptoms were "dizziness", "pale" and "no scar". "Murder with venomous snake" or "Venom killing" is inferred as a feasible measure for Roylott. The reason was the whip which was in his room.

**Total Judgement**

From the above, it is inferred that Roylott killed Julia by the use of a venomous snake for money.

### 3.3   Submission of Team FLL-ML, Fujitsu

The FLL-ML team applied machine learning based methods to predicting the criminal of "The Adventure of the Speckled Band" (TSB) and explaining the reason for the prediction.

In order to predict the criminal of TSB, we trained a classifier for classifying TSB characters into 3 classes (Criminal, Victim, Other) by using the word vectors corresponding to characters appearing in the short stories of Sherlock Holmes (SH) series other than TSB. We automatically extracted 14,619 word segmented sentences and the list of SH characters from 22 short stories of SH in the Aozora Bunko as training data for the criminal prediction model. We also manually annotated each character in the list with the 3-class classification.

As the reason for judging as the criminal, we presented the sentences which indicated the motive/means to cause the crime from the sentences which were semantically close to the criminal. We trained another classifier for classifying sentences into 3 classes (Motive, Means, Other) with 2,930 annotated sentences from the short stories of SH other than TSB and then the average word vectors in the sentences (sentence vectors) were used for the sentence features. Semantic distances between the criminal and sentences are calculated by the Euclidean distances between the word vector corresponding to the criminal and sentence vectors. As for the output of the actual sentence indicating motives/means, top 30 sentences that have high motives/means scores were presented from sentences close to criminals as candidates, and sentences which indicate motives/means were manually selected from the candidates.

As the result of criminal prediction, the character with the highest criminal score was Roylott, and the next was Helen. The victim was expected to be Julia, if Roylott was the criminal. Roylott's motive for the crime was presumed to be a tantrum or money problem, but the means of the crime could not be identified. Table 1 shows an example of sentences presumed to indicate the motives for the crime.

The result of the criminal prediction was not different from the general interpretation of TSB. As for the motives of Roylott's crime, tantrums and money problems were extracted, but because Roylott's crime was premeditated, a tantrum was inappropriate as a motive. However, since there are no other teams that consider tantrums as a motive in this Knowledge Graph Reasoning Challenge, we have found that machine learning can be used to roughly grasp matters that are difficult to cover by knowledge. On the other hand, our explanation method is too simple for explaining complicated procedure such as the means of the crime in TSB. As the future, we have to consider the construction of the knowledge which can explain complicated procedure and how to associate the knowledge with the prediction results.

**Table 1.** Example of sentences presumed to indicate the motives

| Motive | Basis |
| --- | --- |
| Tantrum | In a fit of anger, however, caused by some robberies which had been perpetrated in the house, he beat his native butler to death and narrowly escaped a capital sentence. |
| Money | Nothing was left save a few acres of ground, and the two-hundred-yearold house, which is itself crushed under a heavy mortgage. |

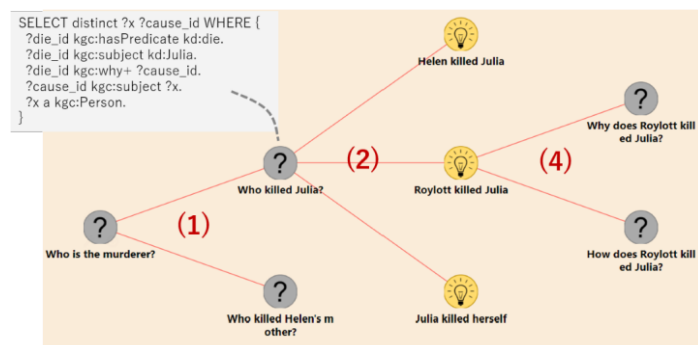

**Fig.6.** An instance of IBIS structure

### 3.4    Submission of Nagoya Institute of Technology

In this section, we describe an approach to the explainability using agent-based discussion on "who is the criminal", which was conceived in Nagoya Institute of Technology. This unimplemented idea is based on the assumption that structured discussion as questions and answers can improve the explainability.

In mystery novels, sidekick characters accompanying detectives have an important role. The sidekicks often ask questions to the detecteves and try to make naive inferences or hypotheses. Such scenes with discussion enable readers to grasp the background of the incident and the process of the inference. From this viewpoint, the explainability can be ensured by generating questions and multiple hypotheses for each question and by evaluating each hypothesis. This idea is inspired by a research project [5,6] to develop autonomous facilitator agents for online discussion based on generating facilitators' questions using the issue-based information system (IBIS) [7].

The procedure of the automated agent-based discussion is designed as follows, where $G$ is the knowledge graph representing the content of the novel.

1. As shown in (1) of Figure 6, issue nodes for asking who is the murderer for each victim $v$ appearing in $G$ are appended to the IBIS structure.
2. A hypothesis isKilledBy($v,x$), i.e., $v$ is killed by $x$, is generated for each pair of a victim $v$ and a murderer $x$. As shown in (2) of Figure 6, the hypothesis isKilledBy($v,x$) is appended to the IBIS structure as an idea node.
3. A discussion agent $d(v,x)$ is assigned to each hypothesis isKilledBy($v,x$) to try generate detail explanation of the hypothesis. The agent $d(v,x)$ firstly has a knowledge graph $G_{v,x}$, a duplication of $G$. The agent $d(v,x)$ appends isKilledBy($v,x$) to $G_{v,x}$ and the IBIS node.
4. A facilitator agent generate questions such as "How does $x$ killed $v$?" and "Why does $x$ killed $v$?" and append them to the IBIS structure as issue nodes as shown in (4) of Figure 6.
5. A discussion agent $d(v,x)$ respectively try to generate hypotheses to answer the questions from the facilitator agent, e.g., a hypothesis $how(v,x)$ about how $x$ killed $v$ and a hypothesis $why(v,x)$ about why $x$ killed $v$. These explanations are appended to $G_{v,x}$ and the IBIS structure.
6. A discussion agent $d(v,x)$ respectively try to disprove a hypothesis isKilledBy($v,x'$) for each $x' \neq x$. $d(v,x)$ tries to generate counterargument against

    $how(v,x')$ and $why(v,x')$. If the counterargument is successfully generated, $d(v,x)$ appends it to $G_{v,x'}$, the knowledge of $d(v,x')$, and to the IBIS structure.

7. The facilitator agent evaluates each $cnsstcy(v,x)$, scores representing the consistency of $G_{v,x}$ including the hypothesis $isKilledBy(v,x)$, and selects a candidate of murderer $x_v = \arg\max_x \{cnsstcy(v,x)\}$ for each victim $v$. The selected discussion agent $d(v,x_v)$ outputs the hypothesis $isKilledBy(v,x_v)$ and its explanation $how(v,x_v)$ and $why(v,x_v)$.

In this procedure, there are some remaining issues on developing a generation method of questions and hypotheses and on defining the consistency metric $cnsstcy(v,x)$. As a future work, the procedure should be improved in respect of these remaining issue and it should be implemented for verification.

## 4  Evaluation

Designing appropriate metrics is necessary for evaluating estimation and reasoning techniques that have explainability. In addition to leading to the correct answer, several metrics, such as explainability, utility, novelty, and performance, should be designed. Then, the proposed approaches are evaluated for their advantages and disadvantages based on the metrics, and classified into categories that correspond to practical use cases. The evaluation is based not only on numerical metrics, but also on a qualitative comparison of the approaches and the common recognition of problems through discussion and peer reviews of evaluators and applicants. The Defense Advanced Research Projects Agency eXplainable AI (DARPA XAI) described in Section 5 states that the current AI techniques have a trade-off between accuracy and explainability, so both properties should be measured. In particular, to measure the effectiveness of the explainability, DARPA XAI rates user satisfaction regarding its clarity and utility. Referring to such activities, we designed the following metrics for this challenge and will further improve them for future challenges. We first share the basic information of the proposed approaches, and then discuss the evaluation of experts and of the general public.

### 4.1  Basic information

First, the following information was investigated and shared with experts in advance. The experts were seven board members of the Special Interest Group on Semantic Web and Ontology (SIGSWO) in the Japan Society of Artificial Intelligence.

**Correctness of the answer:** Check if the resulting criminal was correct or not, regardless of the approach. The criminal, in this case, is the one designated in the novel or story. In the case that several criminals are presented, if the criminal in the novel is included among them, we decided the approach as correct but made a note.

**Feasibilitt of the program:** Check if the submitted program correctly worked and the results were reproduced (excluding idea-only submissions). **Performance of the program:** Referential information on the system environment and performance of the submitted program, except for the idea only. **Amount of data/knowledge to be used:** How much did the approach use the knowledge graph (the total number of scene IDs used)? If the approach used external knowledge and data, we noted information about them.

### 4.2 Expert evaluation

Over more than a week, the experts evaluated the following aspects according to five grades (1–5). For estimation and/or reasoning methods, they considered:
**Significance:** Novelty and technical improvement of the method.
**Applicability:** Is the approach applicable to the other problems? As a guide, 3 means applicable to the other novels and stories and 5 means applicable to other domains.
**Extensibility:** Is the approach expected to have a further technical extension? For example, if a problem is solved, can the process or result be further improved? For use of knowledge and data, they considered the following: **Originality of knowledge/data construction:** Originality of knowledge/data construction (amount × quality × process). For example, how much external knowledge and data were prepared?
**Originality of knowledge/data use:** How efficiently were the provided knowledge and self-constructed knowledge used? For example, was a small set of knowledge used efficiently, or was a large set of knowledge used to simplify the process. They also considered the following:
**Feasibility of idea (for idea only):** Feasibility of idea including algorithms and data/knowledge construction.
**Logical explainability:** Is an explanation logically persuadable? As a guide, 1 indicates no explanation and evidence, 3 indicates that some evidence in any form is provided, and 5 indicates that there is an explanation that is consistent with the estimation and reasoning process.
**Effort:** Amount of effort required for the submission (knowledge/data/system).

**Table 2.** Results per metrics by experts (ave.)

| Application # | A | B | C |
|---|---|---|---|
| Significance | 3.43 | 3.29 | 4.1 |
| Applicability | 2.86 | 3.50 | 3.0 |
| Extensibility | 3.14 | 3.67 | 4.0 |
| Originality of knowledge/data construction | 3.14 | 4.00 | 4.1 |
| Originality of knowledge/data use | 3.43 | 3.17 | 4.1 |
| Feasibility of idea (for idea only) | | | |
| Logical explainability | 3.57 | 2.33 | **4.7** |
| Effort | 3.57 | 4.17 | 4.5 |
| Average | 3.31 | 3.45 | 4.1 |

**Table 3.** Results by experts and the public

| Metrics | Tot. score | | Explain. | |
|---|---|---|---|---|
| | 1st | 2nd | 1st | 2nd |
| Ave. (pub) | 4.26 | 4.04 | 4.11 | 3.97 |
| Med. (pub) | 4.50 | 4.00 | 4.00 | 4.00 |
| S.D. (pub) | 0.67 | 0.65 | 0.78 | 0.69 |
| Ave. (exp) | 4.22 | 4.10 | 4.00 | 4.71 |
| Med. (exp) | – | – | 4.00 | 5.00 |

### 4.3 General examination

Although the experts had a long time to determine whether a logical explanation could be held, the general examination that has a time constraint focused on the psychological aspect of the explanations, that is, the satisfaction with the explanation. In the presentation before the examination, the applicants had 15 min to present their submissions (10 min for idea-only submissions). In November 2018, the 45 participants of the SIGSWO meeting answered to the total score and explainability according to five grades (1–5). The forms were distributed in advance, and explained. After all the presentations, the forms were collected.

We added the total score to include psychological impressions other than explainability, such as presentation quality and entertainment aspects. If we only had a score for explainability, such aspects could be mixed in the explainability score.

### 4.4 Evaluation results

Tables 2 and 3 present part of the evaluation results. In terms of the results of the general examination, we compared averages, medians, and standard deviations of the total scores and the scores for the explainability. Comparing the first and second prizes, we found that the averages of both the total score and the score for the explainability were higher for this first prize. The median of the total score in the first prize was higher than that of the second prize, but the median of the score for the explainability in the first prize was the same as the that of the second prize. Moreover, the standard deviations of both the total score and the score for the explainability were bigger for the first prize than for the second prize. The paired

*t*-test ($\alpha$ = 0.05) indicated that difference in total score had a statistically significant difference between the first and second prizes, but the score for the explainability was not significantly different.

In terms of the results of the experts, the averages of each metric in the first prize were higher than those of the second prize, except for the explainability score, which was statistically significantly higher for the second prize according to the *t*-test. We should note that the standard deviations of the averages for each metric were less than 0.1; thus, there were no big differences among their evaluations. Among the metrics, explainability had the least variance, and the effort required had the biggest variance.

Therefore, the final decision was left to the expert peer review. As a result, we decided this prize order, since the metrics other than the explainability of the first prize were higher than or equal to the second prize. At the same time, the evaluation of the estimation and reasoning techniques including explainability, which was a key goal of this challenge, was left to the future challenge. In addition to the first and second prize, we gave a best resource and a best idea prize based on the comments of the experts.

## 5  Related Work

In terms of AI development with explainability, the Defense Advanced Research Projects Agency (DARPA) started the eXplainable AI (XAI) project in 2017. DARPA XAI is a research and development project to help soldiers understand, trust, and manage future AI partners[3], and it is developing machine learning techniques to generate more explainable models while retaining the high-level learning function. At the same time, the model should be able to translate an explanation that is more understandable and useful to human users using the latest human-computer interaction (HI) techniques. The integration of the eXplainable AI model and human interaction was intended from the beginning of the project. Specifically, two tasks corresponding to the DARPA missions, data analytics, and autonomy were set as problems to be solved. The data analytics task is technically a classification problem of multimedia data and indicates the basis for the decision to the human analyst when automatically identifying targets from images. The autonomy task is a reinforcement learning problem of an autonomous system, such as the type used in drones and robots, and presents why the next action was selected in a given situation to human operators using the autopilot mode. To indicate the reason, three methods are discussed. Deep explanation shows which features are important for identification in deep learning [2]. Interpretable models mainly use random forests, Bayesian networks, and probabilistic logics, and they show the meanings and correlations of nodes in the constructed network. Model induction handles a model as a black box and creates a simpler and more analytical model with the same input and output.

---

[3] https://www.darpa.mil/program/explainable-artificial-intelligence

The explainability of AI also has a social need. The Japanese Ministry of Internal Affairs and Communications prepared ten general principles for AI promotion and its risk reduction in 2018. Although these are not rules, they are expected to evoke public opinion by discussion in and outside Japan. The principle of transparency (#9) defines that service providers and business users of AI must pay attention to the verification possibilities of input and output, and the explainability of AI system/service results. The principle of accountability (#10) defines that service providers and business users of AI should have accountability to stakeholders including consumers and end-users. In the European Union (EU), article 22 in the General Data Protection Regulation (GD-PR) enforced May 2018 defines that service providers of data-based decision-making must have the responsibility to safeguard users rights, at least the right to obtain human intervention.

Accordingly, in top conferences of AI and neural networks, such as IJCAI, AAAI, NIPS, and ICML, papers and workshops that have "expandability" as a keyword and that analyze the properties of AI models have significantly increased since 2016. However, there is no research activity like this challenge, which uses knowledge graphs including social problems as common test-sets and tries to solve the problems with explainability, aiming to integrate inductive estimation and deductive reasoning.

Although knowledge graphs and schema constructed for this challenge are our original work, related works include EventKG[8], ECG[9], and Drammer[10]. Knowledge graphs such as Wikidata and DBpedia focus on entities of persons and objects, but EventKG is a Knowles graph that describes 690,000 historic and modern events to generate question answering and history (timeline) from specific aspects. It uses a schema that extends temporal relation expressions based on the Simple Event Model[11]. Although there are similarities to our schema, e.g., definitions of event relationships, the granularity of their events is much bigger than one of our scenes; thus, it is difficult to describe who, whom, and what for each scene using the EventKG schema [4]. ECG provides a schema to annotate extracted information when directly constructing a knowledge graph from a news event that is described in natural languages. However, since it is for automatic extraction, the schema is simple and only includes who, what, where, and when. Drammer focuses on fictional contents and aims not only to sequentially express the content, but also to dramatically present the narrative contents, and it define a schema (or, ontology) including conflict of characters, story segmentation, emotional expression, and belief. That is an intensive work constructed after analysis of several dramas, but is different from our schema for expressing facts and relations in real society.

---

[4] More precisely, it is possible in the EventKG schema, but the graph becomes complicated and is hard to construct such that specific information is difficult to find in the graph.

## 6   Conclusion and future work

This paper reports on the first knowledge graph reasoning challenge, held in 2018, which aimed to promote AI techniques with explainability. In particular, for the development of AI techniques integrating inductive machine learning (estimation) and deductive knowledge reasoning, this challenge constructed and published a knowledge graph based on a short mystery story as an evaluation test-set, collected proposals of methods for estimating and/or reasoning about a criminal with the explanation, and evaluated the proposed approaches based on several metrics.

In the second challenge that started in June 2019, five knowledge graphs were constructed from five different mystery novels. In the third or fourth challenge, we plan to have knowledge graphs of real social problems, e.g., books listing best practices of social problem solving, instead of mystery novels. Moreover, we will make this challenge an international event and promote it as a common academic problem setting. The results of these challenges will be annually reported and open to the public.

**Acknowledgments**

We would like to express our gratitude to all the participants in the workshops, technical meetings, and other events that have been held so far. This work was supported by JSPS KAKENHI Grant Number 19H04168.